\documentclass[letterpaper, 10 pt, conference]{ieeeconf/ieeeconf}
\IEEEoverridecommandlockouts
\usepackage{graphicx}
\usepackage{times}  % assumes new font selection scheme installed
\usepackage{amsmath} % assumes amsmath package installed
\usepackage{amssymb} % assumes amsmath package installed

\usepackage[english]{babel}

\usepackage{amssymb}
\usepackage{amsmath}
\usepackage{hyperref}
\usepackage{cleveref}
\usepackage{url}
\usepackage{graphicx}
\usepackage{latexsym}
\usepackage{algorithm}
\usepackage{algpseudocode}
\usepackage[binary-units]{siunitx}
\usepackage{subcaption}
\usepackage{todonotes}

\usepackage{newclude}
\usepackage{multicol}
\usepackage{tabularx}
\usepackage{booktabs}
\usepackage{threeparttable}
\let\labelindent\relax
\usepackage{enumitem}
\usepackage{listings}
\usepackage{wrapfig}
\usepackage{rotating}
\usepackage[export]{adjustbox}
\usepackage{makecell}
\usepackage{multicol}
\usepackage{bm}
\usepackage{wrapfig}

\usepackage[acronym]{glossaries}
\newacronym{wsl}{WSL}{Weakly-Supervised Learning}
\newacronym{fmcw}{FMCW}{Frequency-Modulated Continuous-Wave}
\newacronym{ml}{ML}{Machine Learning}
\newacronym{dl}{DL}{Deep Learning}
\newacronym{cnn}{CNN}{Convolutional Neural Network}
\newacronym{vo}{VO}{Visual Odometry}
\newacronym{ins}{INS}{Inertial Navigation System}
\newacronym{gnss}{GNSS}{Global Navigation Satellite System}
\newacronym{gnssins}{GNSS+INS}{Global Navigation Satellite System and Inertial Navigation System}
\newacronym{rsln}{RSL-Net}{Radar-Satellite Localisation Network}
\newacronym[firstplural=degrees of freedom (DoFs)]{dof}{DoF}{Degree of Freedom}
\newacronym{gan}{GAN}{Generative Adversarial Network}
\newacronym{osm}{OSM}{OpenStreetMap}
\newacronym{api}{API}{Application Programming Interface}
\newacronym{stn}{STN}{Spatial Transformer Net}

\glsdisablehyper

\definecolor{CommentDani}{rgb}{0,0,1}
\definecolor{CommentTim}{rgb}{0.2,0.8,0.2}
\definecolor{CommentPaul}{rgb}{0.9,0,0}
\definecolor{CommentReview}{rgb}{0.9,0.6,0.2}

\newcommand{\fl}{3.0cm}

\newcommand{\figsize}{2.0cm}
\newcommand{\figsizeL}{2.1cm}

\crefname{table}{Table}{Tables}
\crefname{figure}{Figure}{Figures}
\crefname{section}{Section}{Sections}

\usepackage[binary-units]{siunitx}
\usepackage{amsmath}
\usepackage{mathtools}
\usepackage{gensymb}
\usepackage{amsfonts}

\newcommand\norm[1]{\left\lVert#1\right\rVert}

\title{RSL-Net: Localising in Satellite Images From a Radar on the Ground}

\author{Tim Y. Tang, Daniele De Martini, Dan Barnes, and Paul Newman% <-this % stops a space
\thanks{The authors are affiliated with the Oxford Robotics Institute, University of Oxford, United Kingdom.
    {\tt\small \{ttang, daniele, dbarnes, pnewman\}@robots.ox.ac.uk}}%
}

\begin{document}
\maketitle

\begin{abstract}
This paper is about localising a vehicle in an overhead image using FMCW radar mounted on a ground vehicle.
FMCW radar offers extraordinary promise and efficacy for vehicle localisation.
It is impervious to all weather types and lighting conditions.
However the complexity of the interactions between millimetre radar wave and the physical environment makes it a challenging domain.
Infrastructure-free large-scale radar-based localisation is in its infancy.
Typically here a map is built and suitable techniques, compatible with the nature of sensor, are brought to bear.
In this work we eschew the need for a radar-based map; instead we simply use an overhead image -- a resource readily available everywhere.
This paper introduces a method that not only naturally deals with the complexity of the signal type but does so in the context of cross modal processing.
\end{abstract}

\section{Introduction}
\begin{figure}[!htbp]
  \centering
  \begin{subfigure}[t]{\fl}
  \includegraphics[width=\fl]{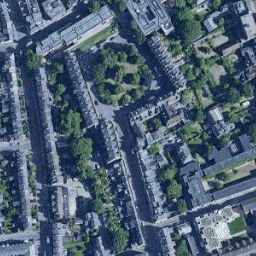}
  \caption{\scriptsize Input satellite image $I_S$} %\vspace{2mm}}
  \end{subfigure}
  \qquad
  \begin{subfigure}[t]{\fl}
  \includegraphics[width=\fl]{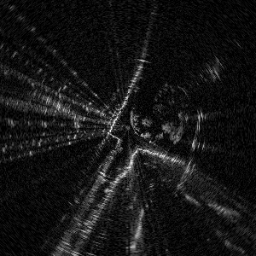}
  \caption{\scriptsize Input radar image $I_R$} %\vspace{2mm}}
  \end{subfigure}
  %
%	\vspace{3mm}
  %
  \begin{subfigure}[t]{\fl}
  \includegraphics[width=\fl]{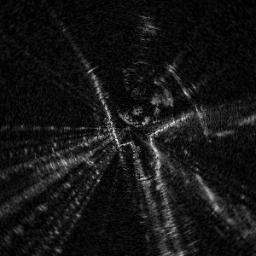}
  \caption{\scriptsize Radar image with rotation offset uncovered, $\hat{I}_{R,*}$}
  \end{subfigure}
  \qquad
  \begin{subfigure}[t]{\fl}
  \includegraphics[width=\fl]{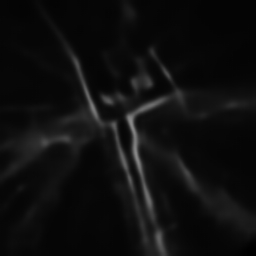}
  \caption{\scriptsize Synthetic radar image $\hat{I}$}
  \end{subfigure}
  %
%  \vspace{3mm}
  %
  \begin{subfigure}[t]{\fl}
  \includegraphics[width=\fl]{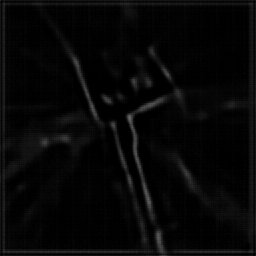}
  \caption{\scriptsize Deep embedding $e_A(\hat{I})$}
  \end{subfigure}
  \qquad
  \begin{subfigure}[t]{\fl}
  \includegraphics[width=\fl]{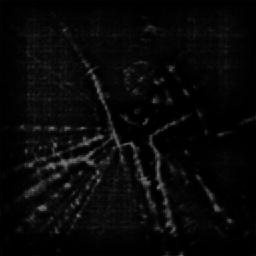}
  \caption{\scriptsize Deep embedding $e_B(\hat{I}_{R,*})$}
  \end{subfigure}
  \caption{\label{fig:RSL_various_stages} \footnotesize Given an input satellite-radar pair $I_S$ and $I_R$ with an unknown offset in $SE(2),$ RSL-Net first uncovers the rotation offset $\theta$ and creates $\hat{I}_{R,*},$ which is $I_R$ rotated to match the heading as in $I_S.$
Taking in $I_S$ and $\hat{I}_{R,*}$ as inputs, a generator network produces $\hat{I}$ conditioned on the appearance of $\hat{I}_{R,*}$ and the pose of $I_S.$
The images $\hat{I}$ and $\hat{I}_{R,*}$ are projected to a joint deep embedding by functions $e_A$ and $e_B,$ where their translation offset in $(x,y)$ is found.
}%
\vspace{-4mm}
\end{figure}

For autonomous vehicles to operate reliably they must be able to localise with respect to their surrounding environment.
While localisation in benign environments is generally a well-studied concern, long-term localisation in complex environments under changing weather and scene conditions remains an open problem.
Localisation using vision sensors can suffer due to a different lighting condition as in the map, while, although robust against lighting conditions, lidars are sensitive to foul weather such as rain, fog, and snow.

Current approaches typically involve mapping the environment multiple times using on-board sensors across different scene and appearance conditions \cite{churchill2013experience,pomerleau2014long}, or learning an appearance transfer across domains, such as daytime and night images \cite{porav2018adversarial}.
For scenarios of unusable or missing prior sensor maps, publicly available overhead imagery offers a cheap, easily accessible fallback solution.

To this end, we propose \gls{rsln}, which localises an on-board \gls{fmcw} scanning radar against overhead imagery.
Scanning radar sensors are invariant to lighting and weather, making them ideal for operating outdoor, and radar images can be displayed in a ``birds-eye'' view similar to overhead imagery, making them suitable for the task.
Though we make use of Google satellite imagery in this work, \gls{rsln} is not limited to satellite images and can be trained for any types of overhead imagery.

Given as inputs an overhead image and a radar image with an unknown $SE(2)$ pose offset and sufficient overlap, \gls{rsln} generates a synthetic image that preserves the appearance of the input radar image and the pose of the input overhead image.
Shown in \cref{fig:RSL_various_stages}, this is done by first uncovering the rotation offset between the two (c), and then generating the synthetic image (d).
The synthetic image and radar image are then embedded into a joint domain (e and f), where the 2D position offset is found by an exhaustive search in the $(x, y)$ solution space, thus solving the remaining two \glspl{dof} of the $SE(2)$ pose.

We do not seek to solve the general place recognition problem, but rather assume that a very coarse initial pose estimate can be found by other means, for example a combination of radar odometry \cite{cen2018precise} and noisy \gls{gnss}.
Once we have a coarse initial pose estimate (10s of metres error is expected), our method utilises overhead imagery to compute a refined metric localisation.

Despite radars being immune to weather and lighting conditions, localising a ground range sensor against overhead imagery remains challenging for several reasons.
Radar images are collections of frequency vs energy tuples (where frequency is a proxy for range), and are therefore intrinsically different than overhead images, making a direct comparison quite daunting.
As is true for most on-board sensors, due to occlusions, the radar only observes a limited region of the scene captured by the overhead image.
Parts of the scene such as cars, pedestrians, and temporary road signs may exist in the live data but not in overhead imagery, or vice versa.

To the best of our knowledge, this paper presents the first method to learn the metric localisation of an on-board range sensor using overhead imagery.
By evaluating across datasets containing both an urban setting and highway driving in rural regions, we show that our method achieves high localisation accuracy in various types of environments.
We validate on test data covering approximately \SI{10}{\minute} of driving and \SI{10}{\kilo\metre} of distance travelled.

\section{Related Work}
\label{section:related_works}

\subsection{Localisation Against Overhead Imagery} 
Prior work for metrically localising a ground sensor against overhead imagery are mostly classical estimation methods.
Noda et al. \cite{noda2010vehicle} localised a ground camera against overhead images using SURF features. 
The methods in \cite{leung2008localization, pink2008visual, viswanathan2014vision} achieve a similar task using other hand-crafted features. 
Senlet et al. \cite{senlet2011framework} presented a framework for dense matching between ground stereo images and satellite images. 
The work in \cite{kummerle2011large} localises a ground laser using edge points detected from aerial images.
The method in \cite{de2015re} registers lidar intensity images against overhead images by maximising mutual information.
By finding features from vertical structures, FLAG \cite{wang2017flag} achieves air-ground outdoor localisation as well as indoor localisation using floor plans.
Forster et al. \cite{forster2013air} localised a ground laser using dense reconstruction from overhead images taken by an aerial vehicle.
These methods rely on hand-crafted systems to work in specific setups, but does not demonstrate generalisation in previously unseen environments.

Learning-based methods have been proposed to solve place recognition using overhead imagery, but not metric localisation.
Chu et al. \cite{chu2015accurate} learned a feature dictionary to compute the probability that a ground image is taken with a particular pose using overhead imagery.
Given a database of satellite images, Kim et al. \cite{kim2017satellite} trained a Siamese network to learn matched ground-satellite image pairs, which are used in particle filter localisation.
These methods do not learn the sensor pose end-to-end.
Similarly, work in ground-to-aerial geolocalisation such as \cite{lin2015learning, workman2015wide, hu2018cvm} addresses place recognition by finding ground-satellite image matches, but does not learn metric localisation of the ground sensor.

\subsection{Localisation Using Other Publicly Available Data} 
Some methods address the metric localisation of a ground sensor using other types of off-the-shelf, publicly available data, particularly \gls{osm}.  
Hentschel et al. \cite{hentschel2010autonomous} used building tags in \gls{osm} as additional observations to update the weights of particles in particle filter localisation.
Brubaker et al. \cite{brubaker2013lost} built a graph-based representation of \gls{osm} and a probabilistic model to match \gls{vo} paths to the map.
Similarly, Floros et al. \cite{floros2013openstreetslam} used Chamfer matching to match \gls{vo} paths to \gls{osm} paths. Methods utilising \gls{osm} typically require labels for buildings or path information, while our method relies on overhead imagery and does not depend on labelled information.

\subsection{Deep Learning for Outdoor Localisation} 
Given training data from a previously seen environment, the work by Kendall et al. \cite{kendall2015posenet, kendall2017geometric} directly regresses the 6-\gls{dof} camera pose for relocalisation.
For a live query image, Laskar et al. \cite{laskar2017camera} learned a descriptor to find matches from a database of prior images, and regressed the pairwise relative pose.
The method presented in \cite{brachmann2018learning} solves camera relocalisation by learning 3D scene coordinates.
MapNet \cite{brahmbhatt2018geometry} learns a data-driven map representation for localisation, where geometric constraints used for state estimation are formulated as network losses.
Sarlin et al. \cite{sarlin2019coarse} learned keypoints and descriptors for camera localisation.
Recent work in \cite{lu2019l3} addresses localisation using lidar point-clouds by learning keypoint descriptors through 3D convolutions.

%Related to our work is the method by Barsan et al. \cite{barsan2018learning}, where live lidar intensity images and lidar intensity maps are embedded to a joint space for localisation.
%Inspired by \cite{barsan2018learning}, our method also searches for the translational offset between two deep embeddings by convolving them.
%Our work and the methods in \cite{lu2019l3} and \cite{barsan2018learning} compute the pose offset by exhaustively searching through the solution space, rather than regressing the pose with a fully connected layer as in \cite{kendall2015posenet} and \cite{kendall2017geometric}.
%Existing work relies on prior data gathered with the same sensor modality as the live data, whereas our method is concerned with localising a radar against overhead imagery.

Related to our work is the method by Barnes et al. \cite{barnes2019masking}, where radar images are projected to a deep embedding for pose estimation.
Inspired by \cite{barnes2019masking} and work by Barsan et al. \cite{barsan2018learning} which uses a similar approach on lidar intensity images, our method also searches for the translational offset between two deep embeddings by convolving them.
Our work and the methods in \cite{lu2019l3}, \cite{barnes2019masking} and \cite{barsan2018learning} compute the pose by exhaustively searching through the solution space, rather than regressing the pose with a fully connected layer as in \cite{kendall2015posenet} and \cite{kendall2017geometric}.
Existing work relies on prior data gathered with the same sensor modality as the live data, whereas our method is concerned with localising a radar against overhead imagery.

\subsection{Conditional Image Generation} 
An important part of \gls{rsln} is a generator network conditioned on an input overhead image and an input radar image. 
Building upon Pix2Pix \cite{isola2017image} which performs image-to-image translation using conditional \glspl{gan}, the work in \cite{zhu2017toward} achieves multi-modal generation where the input image is conditioned on a random latent vector.
Similarly, IcGAN \cite{perarnau2016invertible} edits an input image using a conditional vector.
The network in \cite{lin2018conditional} takes in two input images, where the output contains attributes from both inputs, for example shape as in the first input and color as in the second input.
The generator used in \cite{ma2017pose} takes in a person image and a pose image as inputs; the network generates an output person image that preserves the appearance of the input person image and the pose in the pose image.
Likewise, the generator we use takes in an overhead image and a radar image as inputs, and outputs a generated image that preserves the appearance of the input radar image, with a pose being that of the overhead image.

\section{Radar-Satellite Localisation Network}%
\label{section:method}
Given a live radar image $I_R$ and a satellite image $I_S,$ sampled at a coarse initial pose estimate $(x_0, y_0, \theta_0),$
%for example from a combination of odometry and noisy \gls{gnss}, 
we seek to estimate the radar's pose by localising against satellite imagery. 
We stress that the initial pose estimate does not need to be accurate, but only needs to be good enough such that a sufficient portion of the scene captured by the satellite image is also observed by the radar.
The scale of the satellite image is known a priori, the only unknown pose parameters are a rotation and a translation, making this an estimation problem in $SE(2).$
$I_R$ and $I_S$ are fed into \gls{rsln} to compute the $SE(2)$ offset between them parametrised as $(x, y, \theta),$ which provides a refined solution to the radar's pose. 

\subsection{Overview of the Pipeline}
To localise a radar image against a satellite image, our approach first seeks to generate a synthetic radar image from the satellite image, which is used downstream to compute the pose against the radar image itself.
Specifically, we wish to learn a function $f$ such that 
\begin{equation}
\label{eq:func_f}
f : (I_R, I_S) \rightarrow I,
\end{equation}
where $I$ is a generated radar image that preserves the appearance of $I_R$ but has the same pose as $I_S.$
We can obtain the targets for supervising the training as we have the ground truth pose between $I_R$ and $I_S,$ which we use to transform $I_R$ into $I.$

We learn $f(I_R, I_S)$ rather than $f(I_S)$ since radar data is available at inference to condition $f.$
We show in \cref{sec:method_comparison} that this leads to greatly improved results than learning $f: I_S \rightarrow I,$ which does not depend on $I_R.$
However, training $f$ to condition on both $I_R$ and $I_S$ is non-trivial as $I_R$ and $I_S$ are offset by an $SE(2)$ pose.
Standard \glspl{cnn} are equivariant\footnote{A mapping $f:\mathcal{X} \rightarrow \mathcal{Y}$ is \textit{equivariant} to a group of transformations $\Phi,$ if for any $\phi \in \Phi,$ there exists a transformation $\psi \in \Psi$ such that $\psi\big( f(\chi) \big) = f\big(\phi(\chi)\big), \forall \chi \in \mathcal{X}.$ \textit{Invariance} is a special case of equivariance, such that $f$ is invariant to $\Phi$ if, for any $\phi \in \Phi,$ $f(\chi) = f\big( \phi(\chi) \big).$} to translation, but not to rotation.
Replacing \gls{cnn} filters with a specific family of filters \cite{worrall2017harmonic,weiler2018learning} can allow for equivariance to other transformations, but limits the types of filters the network learns.
Other approaches learn canonical representations \cite{Esteves2018PolarTN}, which are not applicable to our task as canonical representations do not exist for arbitrary radar images.

For this reason, rather than tackling the issue with $SE(2)$ equivariance, we break the problem of learning $f$ into two sequential steps.
First, given $I_R$ and $I_S$ with an unknown $SE(2)$ offset, we seek to learn a function $h$ that infers the rotation offset between them; this is done by searching through a discretized solution space of rotations. 
Specifically, we first rotate $I_R$ about its centre by angles with small increments $\{\theta_1, \theta_2, \dots, \theta_{N_\theta}\}$ to create an ensemble of images $\{I_R\}_{N_\theta} = \{I_{R,1}, I_{R,2}, \cdots, I_{R,{N_\theta}}\},$ where each $I_{R,n}$ is a rotated version of $I_R.$ 
We seek to learn
\begin{equation}
\label{eq:func_h}
h: \big( \{I_R\}_{N_\theta}, I_S \big) \rightarrow I_{R, *},
\end{equation}
where $I_{R, *}$ is an image from the ensemble whose heading $\theta$ most closely matches with that of $I_S.$
$I_{R, *}$ is then offset with $I_S$ only in translation, while the other images in the ensemble $\{I_R\}_{N_\theta}$ are offset with $I_S$ by rotation and translation.
The optimal $\theta$ estimate is also inferred in this process.

Without any prior estimate on heading, the search space $\{\theta_1, \theta_2, \dots, \theta_{N_\theta}\}$ should contain rotations ranging from $-\pi$ to $\pi$ with fine discretization. 
However, we can limit the range of solution space to search if we have a coarse estimate on heading, as shown in \cref{sec:experiments}.

After finding $I_{R, *}$ we learn a function $g$ to generate $I:$ 
\begin{equation}
\label{eq:func_g}
g: ({I_{R,*}, I_S}) \rightarrow I.
\end{equation}

Learning $g$ to condition on $I_{R,*}$ and $I_S$ is simpler than learning $f$ to condition on $I_R$ and $I_S,$ as in \eqref{eq:func_f}, since $I_{R,*}$ and $I_S$ are offset only in translation which convolutions are inherently equivariant to.
As such, we have replaced learning $f$ with learning $g: \big(h( \{ I_R \}_{N_\theta}, I_S), I_S \big) \rightarrow I.$
By pre-discovering the rotation, we utilise a simple yet effective strategy to take full advantage of the input radar image when generating $I,$ without imposing any restrictions on the network filters learned.

$I$ has the same pose as $I_S,$ therefore $I$ and $I_{R,*}$ are offset only by a translation.
Finally, we embed $I$ and $I_{R,*}$ to a joint domain to compute their offset in translation:
\begin{equation}
\label{eq:func_pos}
p: \big( e_A(I), e_B(I_{R, *}) \big) \rightarrow \begin{bmatrix}
x \\ y
\end{bmatrix} \in \mathbb{R}^2,
\end{equation}
where $e_A$ and $e_B$ are embedding functions to be learned, and $p$ is a pre-defined function that searches for the $(x, y)$ offset between two images by maximising a correlation surface.

Our network is trained in multiple stages. 
We first pre-train individually a ``rotation selector'' network for learning $h,$ and a generator for learning $g.$
We then train all modules together end-to-end, including networks for learning the embedding functions $e_A$ and $e_B.$
The various stages of training are detailed in \cref{section:stage_1,section:stage_2,section:stage_3}, while the overall architecture of \gls{rsln} is shown in \cref{fig:RSL_architecture}.

\subsection{Rotation Selector Network}%
\label{section:stage_1}

\begin{figure*}[!htbp]
\vspace{3mm}
\centering
\includegraphics[width=0.9\textwidth]{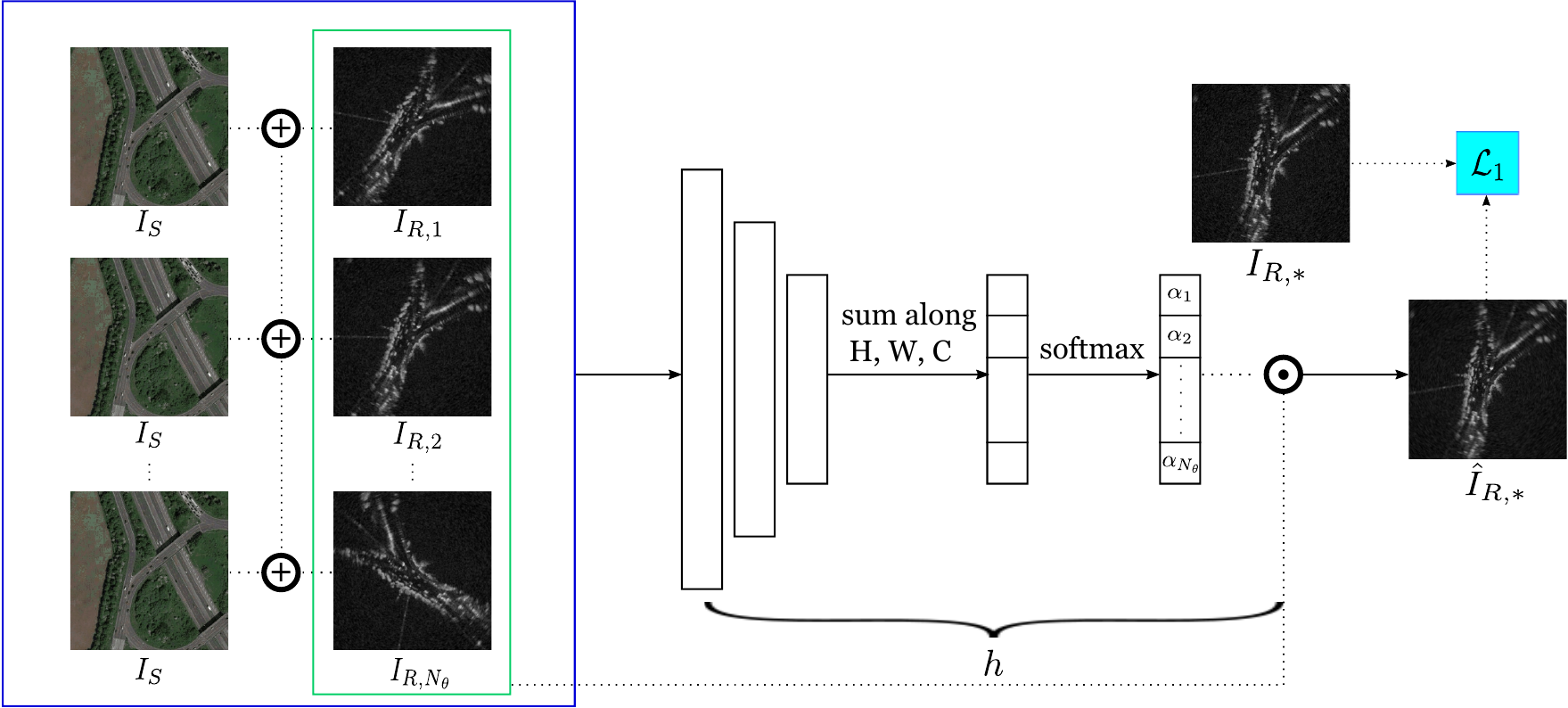}
\caption{\footnotesize \label{fig:rot_selector} Given $I_S,$ we rotate the radar image $I_R$ with $\theta \in \{\theta_1, \dots, \theta_{N_\theta}\}$ to get $\{I_{R,1}, \dots, I_{R,{N_\theta}} \}.$ 
Given the concatenated inputs, the network $h$ outputs a linear combination of the rotated images $I_{R,1}, \dots, I_{R,{N_\theta}},$ where the coefficients are softmaxed. 
The output is a rotated radar image with heading matching that of the input satellite image $I_S.$ We use $\oplus$ to denote concatenation and $\odot$ to denote dot product.
We can obtain the target $I_{R,*}$ by rotating $I_R$ according to the ground truth heading.}
\vspace{-6mm}
\end{figure*}

We design a rotation selector network such that the network learns to select an image $I_{R,*}$ from $\{I_R\}_{N_\theta}$ with a heading angle $\theta$ that most closely matches with $I_S.$
To make the network differentiable, as an approximation of \eqref{eq:func_h}, the output of the network $\hat{I}_{R,*}$ is learned to be a linear combination of all images from the input ensemble:
\begin{multline}
\label{eq:softmax_rot_selector}
\begin{split}
\quad \quad \quad \quad
\hat{I}_{R,*} = h\big( \{I_R\}_{N_\theta}, I_S \big) &= \sum_{n=1}^{N_\theta} \alpha_n I_{R,n} \\
I_{R,n} \in \{I_R\}_{N_\theta}&, \sum_{n=1}^{N_\theta} \alpha_n = 1, \\
\end{split}
\end{multline}
where $\alpha_n$ are learned scalar coefficients.

The architecture of the network is shown in \cref{fig:rot_selector}: the network consists of 5 down-convolutional layers with stride 2 followed by 6 ResNet blocks \cite{he2016deep}, where we sum along the $H, W, C$ dimensions to arrive at a vector $\begin{bmatrix}
\alpha_1 & \cdots & \alpha_{N_\theta}
\end{bmatrix}^T \in \mathbb{R}^{N_\theta}.$
We apply a softmax such that all elements of the vector sum to 1.

The pre-training is supervised using the ground truth image $I_{R,*},$ obtained by rotating $I_R$ to match the heading as in $I_S,$ but maintaining the translational offset.
We use an L1 loss between the output image and the target image:
\begin{equation}
\label{eq:loss_h}
\mathcal{L}_h = \norm{h\big( \{I_R\}_{N_\theta}, I_S \big) - I_{R,*}}_1.
\end{equation}

Ideally, the computed coefficient will be close to 1 for the image whose heading is the nearest to the heading in $I_S,$ and close to 0 for the other images in the ensemble.
The optimal $\theta$ is found by selecting the rotation with the maximum coefficient $\alpha_n.$

\subsection{Conditional Image Genarator}%
\label{section:stage_2}

\begin{figure*}[!htbp]
\vspace{2mm}
\centering
\includegraphics[width=\textwidth]{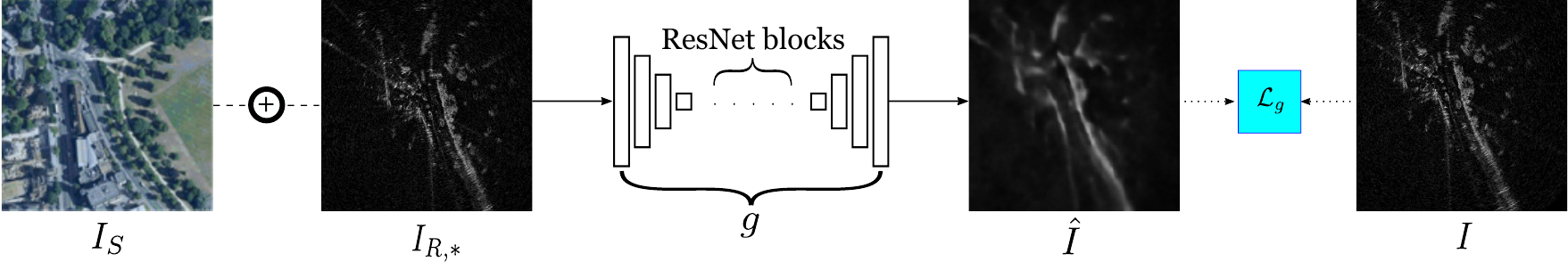}
\captionof{figure}{\footnotesize \label{fig:generator} To pre-train the network $g,$ we rotate $I_R$ to obtain $I_{R,*},$ where the heading of $I_{R,*}$ is the same as $I_S.$ 
There is still an offset in translation between $I_S$ and $I_{R, *}.$ 
The generator conditions on the pose of $I_S$ and the appearance of $I_{R, *},$ and generates $\hat{I}$ which appears like a radar image, and has the same pose as the input satellite image $I_S.$ 
We can obtain the target $I$ by shifting $I_{R,*}$ to match the pose as in $I_S.$}
\vspace{-4mm}
\end{figure*}

The generator is conditioned on input pair $(I_{R, *}, I_S)$ and outputs a generated image $\hat{I}.$
As depicted in \cref{fig:generator}, we use a standard encoder-decoder structure, with 4 down-convolutional layers with stride 2, followed by 9 ResNet blocks at the bottleneck, and 4 up-convolutional layers to recover the same size as the input image.

We pre-train $g$ using an L1 loss on the image as in \eqref{eq:loss_g}. 
Empirically, we found this to be sufficient and we do not need to introduce additional losses such as a \gls{gan} loss.
\begin{equation}
\label{eq:loss_g}
\mathcal{L}_g = \norm{g(I_{R, *}, I_S) - I}_1.
\end{equation}

To pre-train the generator, we feed to the network $I_{R, *}$ rather than $\hat{I}_{R,*}$ outputted by the rotation selector network, since the ground truth $I_{R, *}$ is available at training.
As such, the pre-training of the generator $g$ is independent from the pre-training of $h$ as in \ref{section:stage_1}.

\subsection{Training RSL-Net End-to-End}
\label{section:stage_3}

\begin{figure*}[!htbp]
\includegraphics[width=\textwidth]{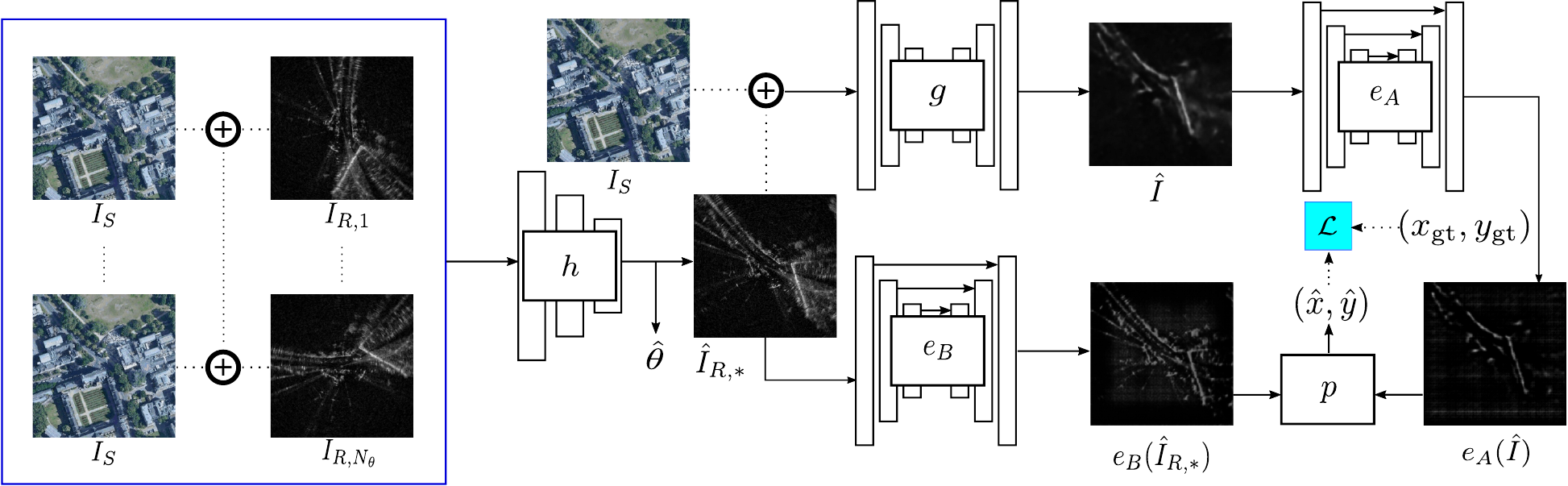}
\caption{\footnotesize \label{fig:RSL_architecture} Having $I_S$ and an ensemble of rotated radar images $\{I_{R,1}, \dots, I_{R,{N_\theta}} \},$ we first infer the rotation by passing through the network $h,$ where we produce $\hat{I}_{R, *}.$
$\hat{I}_{R, *}$ and $I_S$ are passed through $g$ to generate $\hat{I}, $ which has the same pose as $I_S.$
$\hat{I}$ and $\hat{I}_{R, *}$ are projected a joint deep embedding by $e_A$ and $e_B,$ where their translational offset is found.
The only loss term in the final stage is a loss on the computed translation.}
\vspace{-4mm}
\end{figure*}

After pre-training $h$ and $g$ as in \cref{section:stage_1,section:stage_2}, \gls{rsln} is trained end-to-end as illustrated in \cref{fig:RSL_architecture}.
The generator $g$ now takes $\hat{I}_{R,*} = h\big( \{I_R\}_{N_\theta}, I_S \big)$ as input rather than $I_{R,*}.$

The networks $e_A$ and $e_B$ learn to project $\hat{I} = g(\hat{I}_{R,*} , I_S)$ and $\hat{I}_{R,*}$ into a joint embedded domain for pose comparison. 
We use the same U-Net architecture \cite{ronneberger2015u} for  both $e_A$ and $e_B,$ with 6 down-convolutional layers and 6 up-convolutional layers and skip connections between the encoder and the decoder stacks.
The use of U-Nets is motivated by the observation that much of the measurements in $\hat{I}$ and $\hat{I}_{R,*}$ are useful for pose comparison and should be kept.
Regions in the radar image that are distracting for pose comparison, such as returns from reflected signals, or returns on vehicles, are learned to be masked out, as shown in \cref{fig:RSL_various_stages}.

The translational offset is found by shifting $e_A(\hat{I})$ by all $(x, y)$ pairs in the solution space, and finding the one that results in the maximum correlation with $e_B(\hat{I}_{R,*}).$
As such, we compute $(x, y)$ by finding the global maximum of a discretized, 2D correlation surface over a $(x,y)$ solution space of size $N_x \times N_y,$ where $N_x$ and $N_y$ are the number of discretizations in the $x$ and $y$ dimensions.

As noted in \cite{barsan2018learning}, computing correlations across all $(x,y)$ translational shifts in the solution space is equivalent to convolving $e_A(\hat{I})$ with $e_B(\hat{I}_{R,*}).$
Moreover, this operation can be performed very efficiently in the Fourier domain utilising the convolution theorem \cite{barsan2018learning}:
\begin{equation}
e_A * e_B = \mathcal{F}^{-1} \{ \mathcal{F}(e_A) \otimes \mathcal{F}(e_B)\},
\end{equation}
where $*$ denotes convolution, $\otimes$ denotes element-wise multiplication, and $\mathcal{F}$ and $\mathcal{F}^{-1}$ are the Fourier and inverse Fourier transforms.

At this point, the optimal solution $\begin{bmatrix}
\hat{x} & \hat{y}
\end{bmatrix}^T$ is found by applying a softmax over $(x, y)$ across the correlation surface.
We denote this process for computing the translational offset in the embeddings as function $p,$ as in \eqref{eq:func_pos}.

$p$ is differentiable, allowing \gls{rsln} to be trained end-to-end. 
For training end-to-end as in \cref{fig:RSL_architecture}, we supervise using only an L1 loss on the computed translation:
\begin{equation}
\label{eq:loss_pose}
\mathcal{L} = \norm{ p\big(e_A(\hat{I}), e_B(\hat{I}_{R,*}) \big) - \begin{bmatrix}
x_\mathrm{gt} \\ y_\mathrm{gt}
\end{bmatrix} }_1,
\end{equation}
where $x_\mathrm{gt}$ and $y_\mathrm{gt}$ are the ground truth $(x, y)$ shift.

Since $\hat{I}_{R,*} = h\big( \{I_R\}_{N_\theta}, I_S \big)$ and $\hat{I} = g(\hat{I}_{R,*}, I_S),$ \eqref{eq:loss_pose} optimises the networks $h$ and $g$ as well. 
We do not include any additional losses such as $\mathcal{L}_h$ \eqref{eq:loss_h} or $\mathcal{L}_g$ \eqref{eq:loss_g} as in the pre-training stages.
This removes the need for hyperparameter tuning on the relative weights among different loss terms.

In \cite{barsan2018learning} the authors search for the optimal $(x, y, \theta)$ in the deep embeddings from a 3D solution space, where the 2D embedding is rotated $N_\theta$ times with $N_\theta$ being the number of discretizations in the $\theta$ dimension.
In our method, this step is essentially handled previously in the rotation selector network $h,$ where we found the optimal rotation from a discretized solution space $\{\theta_1, \theta_2, \dots, \theta_{N_\theta}\}.$
As such, we only need to search for the optimal translation $(x, y)$ from a 2D solution space in the embedding.

The decision to pre-infer $\theta$ stems from the fact that the rotation needs to be uncovered first in order to condition the generator network $g,$ as the input satellite and radar image to $g$ should only be offset in translation, but not in rotation.
The work in \cite{barsan2018learning} is not concerned with image generation across sensor modalities, and therefore does not have such limitation. 
Regardless, we show in \cref{sec:experiments} that our computation for $\theta$ is accurate despite not performing the search concurrently with $(x,y).$

\section{Dataset and Experimental Setup}
\label{sec:experiments}
%\vspace{-0mm}

Our pipeline is validated using the publicly available Oxford Radar RobotCar Dataset \cite{barnes2019dataset,RobotCarDatasetIJRR}, which features an urban setting, and our own dataset that consists of highway driving in rural regions.

\begin{figure}[!h]
%%\vspace{-2mm}
  \centering
  \begin{subfigure}{3.5cm}{\includegraphics[height=4cm]{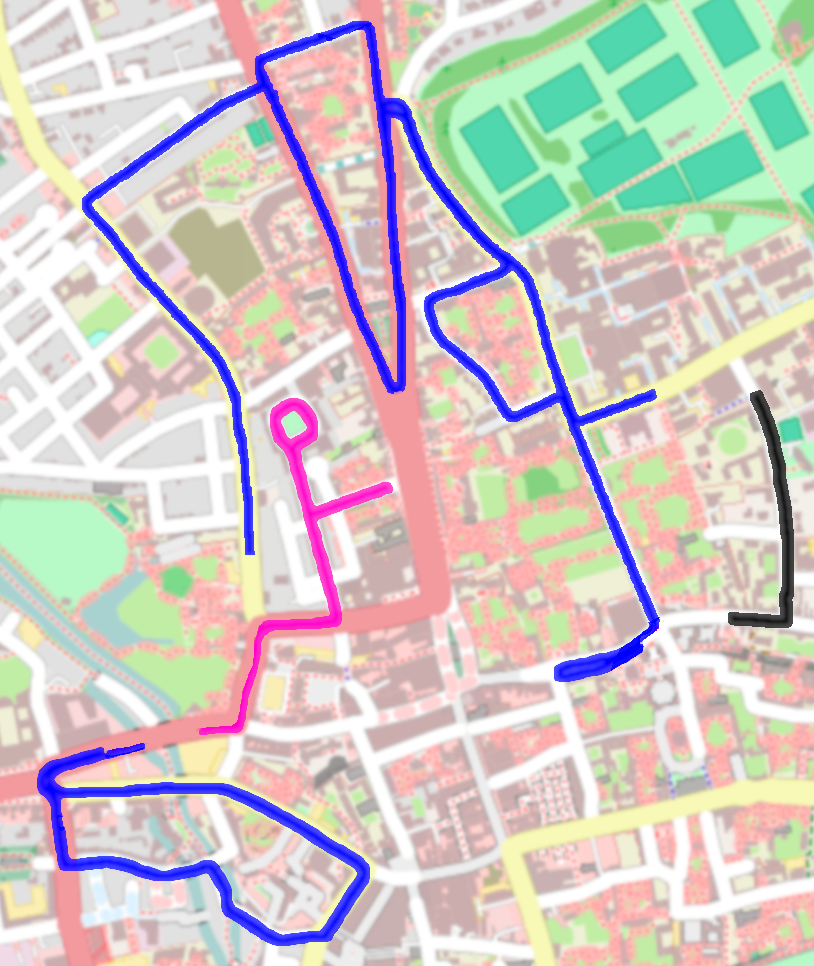} }\end{subfigure}
  \begin{subfigure}{3.5cm}{\includegraphics[height=4cm]{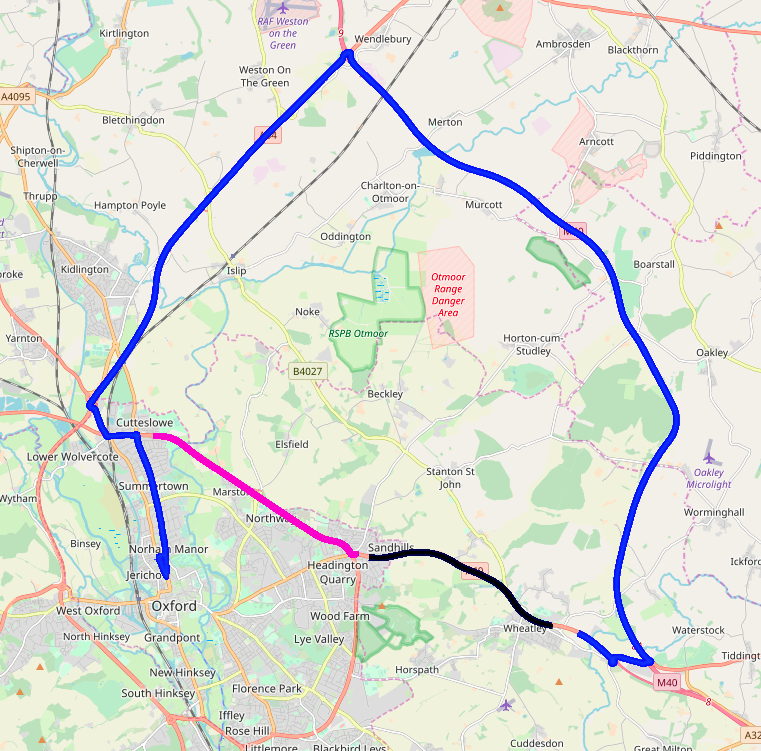} }\end{subfigure}
  \caption{\label{fig:dataset_maps} \footnotesize Routes for the datasets we used are split into training (blue), validation (black), and test (pink), where we discarded data near intersections.}%
  \vspace{-2mm}
\end{figure}

%We used three sequences from the RobotCar Dataset for training, where we split the route into training, validation, and test sets, as shown in \cref{fig:dataset_maps}.
%For simplicity, only the test data from the first sequence was used for evaluation, as all sequences take place on the same route.
%Similarly, we also split the Highway Dataset into training, validation, and test sets, shown in \cref{fig:dataset_maps}.
%We discarded data near the intersections between training, validation, and test data such that the radar does not observe any of the test set scenes in the training and validation data.
%We test on every $5^\mathrm{th}$ frame of the test set routes, where the radar frames arrive at $4~\mathrm{Hz}.$
%This results in 201 test images for the RobotCar Dataset and 261 test samples for the Highway Dataset, spanning a total distance of near $10~\mathrm{km}.$

We used three sequences from the RobotCar Dataset for training, where we split the route into training, validation, and test sets, as shown in \cref{fig:dataset_maps}.
For simplicity, only the test data from the first sequence was used for evaluation, as all sequences take place on the same route.
We also split the Highway Dataset into training, validation, and test sets, shown in \cref{fig:dataset_maps}.
We discarded data near the intersections of the splits such that the radar does not observe any of the test set scenes in the training and validation data.
We test on every $5^\mathrm{th}$ frame of the test set routes, where the radar frames arrive at $4~\mathrm{Hz}.$
This results in 201 test images for the RobotCar Dataset and 261 test samples for the Highway Dataset, spanning a total distance of near $10~\mathrm{km}.$

The RobotCar and Highway datasets are collected with different models of the radar, and exhibit slightly different sensing and noise characteristics.
Despite this, we combine the two datasets together for training, and demonstrate high accuracy on both test sets.
The combined training set consists of 23440 images which spans 98 minutes of driving.
Ground truth latitude, longitude, and heading are available for both datasets with an on-board Novatel \gls{gnss}.
For all of our experiments, we use radar and satellite images with size $256\times 256,$ with a shared resolution of \SI{0.8665}{\si{\metre\per pixel}}.

To simulate a coarse initial position estimate good enough for place recognition but not for metric localisation, we add artificial, uniformly random offsets to the ground truth position when querying for the satellite image, such that the satellite image is off by up to 25 pixels (\SI{21.66}{\metre}) in either $x$ and $y$ directions.
However, we do not limit the solution space for $x$ and $y$ to be 25 pixels, but rather use a search space of $[-128,128]$ pixels for both $x$ and $y.$
The satellite images are queried using Google Maps Static API\footnote{Google Maps Static \gls{api} (2019): \url{https://developers.google.com/maps/documentation/maps-static/}.}.
Similarly, we assume a coarse initial heading estimate is available, such that the solution space for $\theta$ does not need to span $[-\pi, \pi].$
This not only reduces the computational effort needed to search for $\theta,$ but is also crucial against heading ambiguity due to scene symmetry, for example down a straight road where the radar image looks almost invariant after a rotation by $\pi.$ 
In our experiments, we add an artificial, uniformly random offset in the range $\theta_0 \in [\frac{-\pi}{8}, \frac{\pi}{8}]$ to the ground truth heading to simulate a coarse initial estimate, so that the rotation selector network $h$ searches for $\theta$ in the solution space of $[\theta_0 - \frac{\pi}{8}, \theta_0 + \frac{\pi}{8}],$ at intervals of $1\degree.$

\section{Experimental Results}%
\label{sec:results}
%The localisation results on the test data of the RobotCar and Highway datasets are summarised in \cref{tab:results} and \cref{fig:error_dist}.
%Our method achieves a mean error of a few degrees in $\theta.$ 
%We have an error of a few metres in $(x, y)$ on the RobotCar Dataset,  but the translation error is larger on the Highway Dataset.
%This is due to highway settings offering less geometric constraints for pose estimation such as buildings and structures than a urban environment.
%Given a specific threshold $e_\mathrm{th}$ on translation, we compute the ratio of localisations where both $e_x$ and $e_y$ are less than $e_\mathrm{th},$ and $e_\theta$ is less than \SI{5}{\degree}, shown in \cref{fig:ratio_thresh}.
%The average computation time is $99.1~ \mathrm{ms}$ per localisation on a 1080~Ti GPU, allowing us to operate in real-time.

\begin{figure*}[!htbp]
\vspace{2mm}
\centering
\begin{minipage}{.6\textwidth}
  \centering
  \adjincludegraphics[height=7cm,clip]{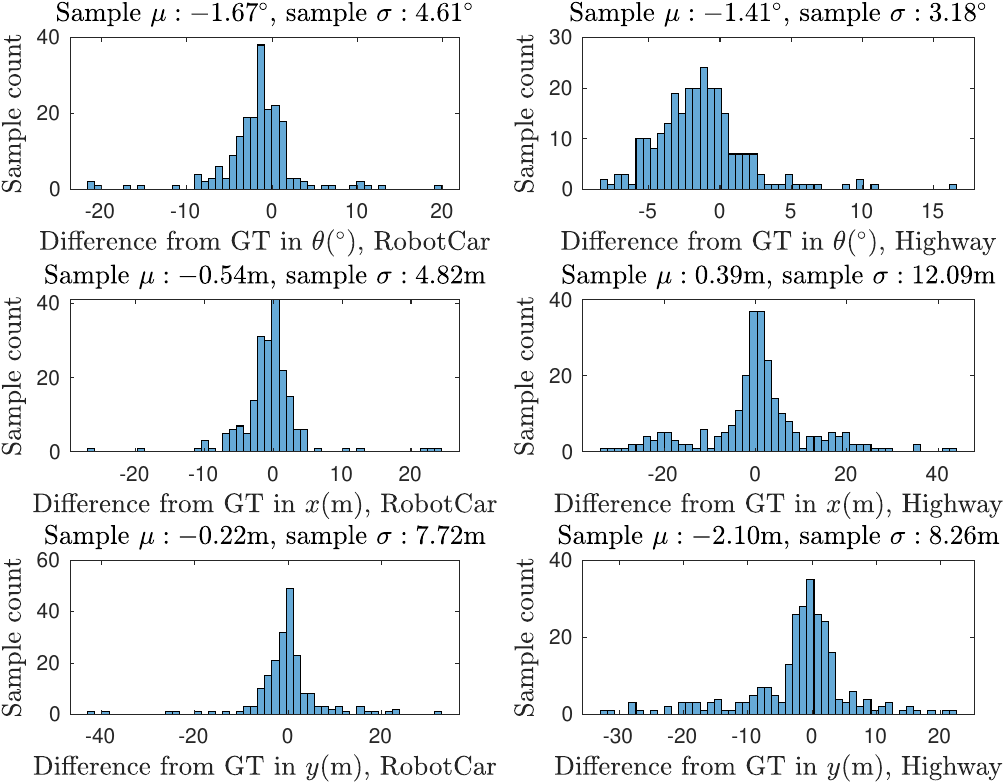}
\captionof{figure}{\footnotesize\label{fig:error_dist} Histogram distribution for errors in $x, y,$ and $\theta$ for the RobotCar and Highway datasets, along with sample mean $\mu$ and standard deviation $\sigma.$} 
\end{minipage}%
\begin{minipage}{.35\textwidth}
  \centering
    \adjincludegraphics[height=4.5cm,clip]{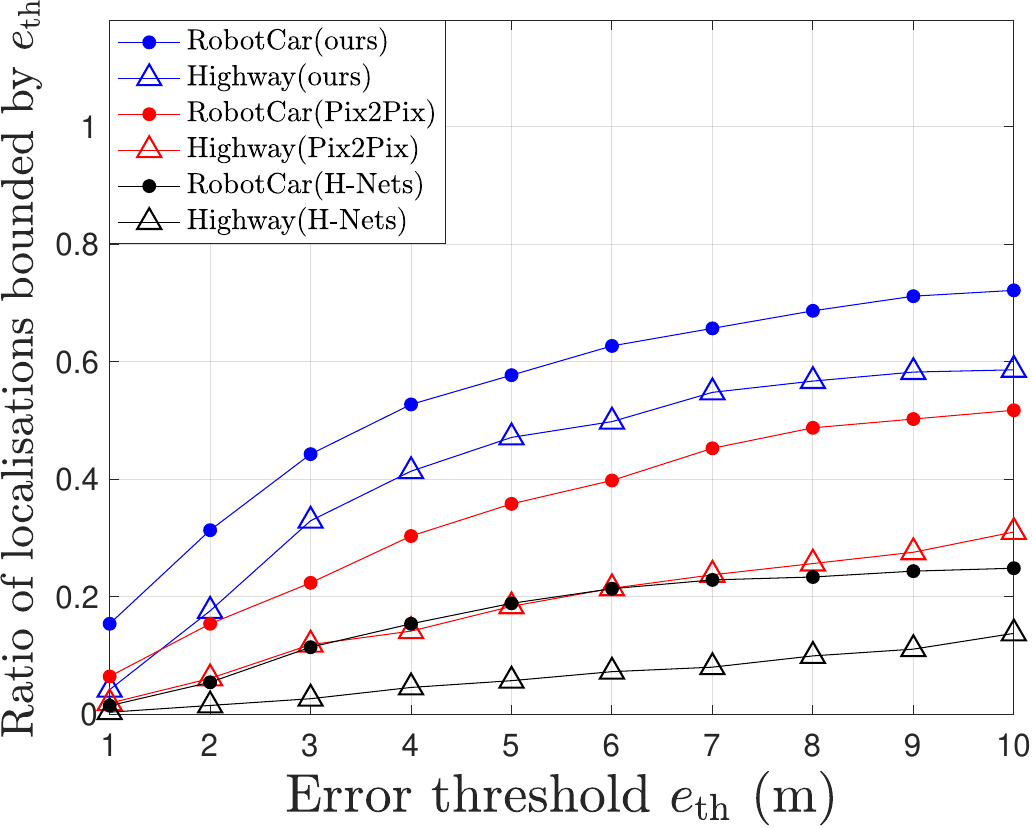}
    \captionof{figure}{\footnotesize\label{fig:ratio_thresh} Ratio of localisations where both $e_x$ and $e_y$ are less than $e_\mathrm{th},$ and $e_\theta$ is less than \SI{5}{\degree}.}
\end{minipage}
\vspace{-4mm}
\end{figure*}

The results on the test data of the RobotCar and Highway datasets are summarised in \cref{tab:results} and \cref{fig:error_dist}.
Our method achieves a mean error of a few degrees in $\theta.$ 
We have an error of a few metres in $(x, y)$ on the RobotCar Dataset,  but the translation error is larger on the Highway Dataset.
This is due to highway settings offering less geometric constraints for pose estimation such as buildings and structures than a urban environment.
Given a specific threshold $e_\mathrm{th}$ on translation, we compute the ratio of localisations where both $e_x$ and $e_y$ are less than $e_\mathrm{th},$ and $e_\theta$ is less than \SI{5}{\degree}, shown in \cref{fig:ratio_thresh}.
The average computation time is $99.1~ \mathrm{ms}$ per localisation on a 1080~Ti GPU, allowing for real-time operation.

\begin{table}[]
\vspace{2mm}
\centering
\footnotesize
\begin{tabular}{c|ccc}
                                          & $e_\theta~(\degree)$ & $e_x~\mathrm{(m)}$ & $e_y~\mathrm{(m)}$ \\
\hline
RobotCar (ours)                    & $\mathbf{3.12}$                & $\mathbf{2.74}$                & $\mathbf{4.26}$\\
Highway (ours)                    & $\mathbf{2.66}$             & $\mathbf{8.13}$                & $\mathbf{5.40}$\\
\hline
RobotCar (Pix2Pix)  & $6.15$                 & $4.07$                 & $6.25$                \\
Highway (Pix2Pix) & $5.98$                 & $8.70$                 & $5.82$                \\
  \hline
 RobotCar (H-Nets) & $4.42$                 & $11.65$                 & $8.74$                \\
Highway (H-Nets) & $6.64$                 & $11.77$                 & $9.14$    \\
\hline
RobotCar (RSL-Net with $g(I_S) \rightarrow I$) &   $3.12$                 &  $10.25$        &  $10.67$              \\
Highway (RSL-Net with $g(I_S) \rightarrow I$) &  $2.66$              &  $10.33$            &  $6.84$            \\
\hline
 RobotCar (RS + regression) & $3.12$                 & $4.10$                 & $5.51$                \\
 Highway (RS + regression) & $2.66$                 & $9.28$                 & $6.67$                \\
 \hline
 RobotCar (RS + STN) & $3.12$                 & $4.35$                 & $4.87$                \\
 Highway (RS + STN) & $2.66$                 & $8.66$                 & $5.87$                \\
  \hline
RobotCar (direct regression) &   $23.56$             & $9.70$           &  $10.87$             \\
 Highway (direct regression) &$24.26$              &  $10.85$               &  $9.43$             \\
    \hline
RobotCar (Canny edges + CV) &  $12.53$               & $17.75$          &  $35.53$      \\
 Highway (Canny edges + CV) &  $10.60$                & $36.73$               &  $42.67$    \\
\end{tabular}
\captionof{table}{\footnotesize \label{tab:results} 
Mean error of \gls{rsln} and other methods on the RobotCar Dataset and Highway Dataset.
$e_x,$ $e_y,$ and $e_\theta$ denotes the mean error in $x,$ $y,$ and $\theta,$ respectively.
RS is abbreviation for rotation selector; CV is abbreviation for cost volume.
The mean error is computed as $e = \frac{\sum \norm{d_i}}{N},$ where $d_i$ is the difference of the $i^\textrm{th}$ sample from ground truth.
We used 201 images in the test set for RobotCar and 261 for Highway.}
\vspace{-7mm}
\end{table}%

%\begin{figure}
% \centering
%  \adjincludegraphics[height=4.5cm,clip]{figs/error_bound_ral}
%\captionof{figure}{\footnotesize\label{fig:ratio_thresh} Ratio of localisations where both $e_x$ and $e_y$ are less than $e_\mathrm{th},$ and $e_\theta$ is less than \SI{5}{\degree}.}
%  \vspace{-2mm}
%\end{figure}

%\begin{figure*}[!htbp]
%\centering
%  \centering
%  \adjincludegraphics[height=10cm,clip]{figs/err_dist}
%  \captionof{figure}{\footnotesize\label{fig:err_dist} abcd.}
%\end{figure*}

%Minipage with vertical error distribution and threshold plots

%\textcolor{blue}{A limitation is that \gls{rsln} may not find a correct heading in places with symmetry in one or more directions.}
%This can be handled by incorporating pose estimates from previous frames for smoothing, or by having a coarse initial heading estimate.

%\begin{figure}[!h]
%  \centering
%  \includegraphics[width=2.5cm]{figs/radar_sym}%
%  \caption{\label{fig:symmetry} \footnotesize An example of scene symmetry.}%
%  \vspace{-4mm}
%\end{figure}

\section{Comparison Against Other Methods} 
\label{sec:method_comparison}
\begin{figure*}[htbp!]%
\vspace{5mm}
\vspace{-1mm}
    \centering
    \begin{subfigure}{2.1cm}
    \centering
    \begin{subfigure}{2cm}{\includegraphics[width=2cm]{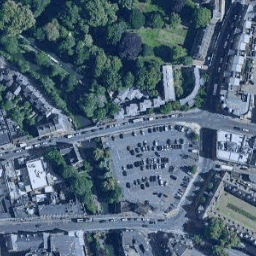}}\end{subfigure}
    \begin{subfigure}{2cm}{\includegraphics[width=2cm]{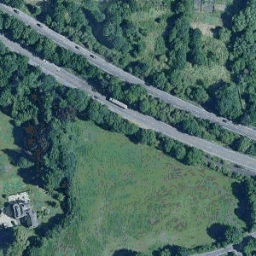}}\end{subfigure}
    \caption{}
    \end{subfigure}
    \begin{subfigure}{2.1cm}
    \centering
    \begin{subfigure}{2cm}{\includegraphics[width=2cm]{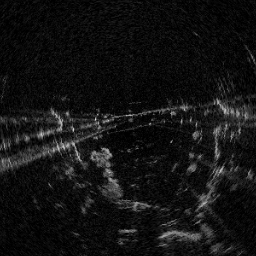}}\end{subfigure}
    \begin{subfigure}{2cm}{\includegraphics[width=2cm]{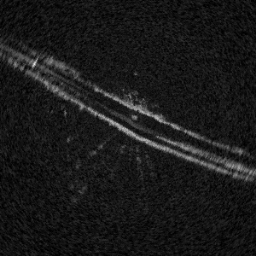}}\end{subfigure}
    \caption{}
    \end{subfigure}
    \begin{subfigure}{2.1cm}
    \centering
    \begin{subfigure}{2cm}{\includegraphics[width=2cm]{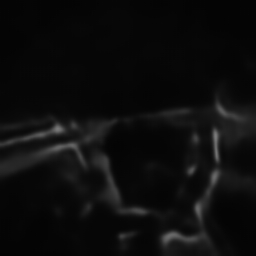}}\end{subfigure}
    \begin{subfigure}{2cm}{\includegraphics[width=2cm]{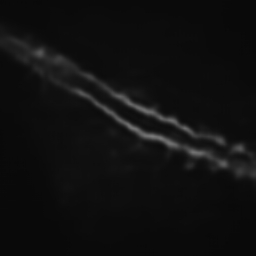}}\end{subfigure}
    \caption{}
    \end{subfigure}
    \begin{subfigure}{2.1cm}
    \centering
    \begin{subfigure}{2cm}{\includegraphics[width=2cm]{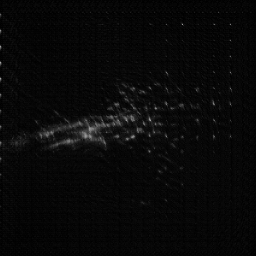}}\end{subfigure}
    \begin{subfigure}{2cm}{\includegraphics[width=2cm]{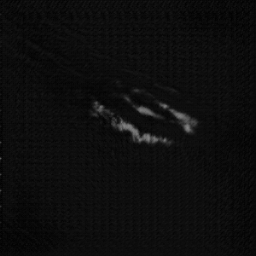}}\end{subfigure}
    \caption{}
    \end{subfigure}
    \begin{subfigure}{2.1cm}
    \centering
    \begin{subfigure}{2cm}{\includegraphics[width=2cm]{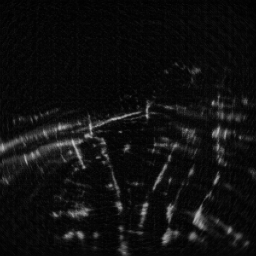}}\end{subfigure}
    \begin{subfigure}{2cm}{\includegraphics[width=2cm]{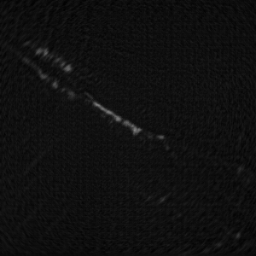}}\end{subfigure}
    \caption{}
    \end{subfigure}
    \begin{subfigure}{2.1cm}
    \centering
    \begin{subfigure}{2cm}{\includegraphics[width=2cm]{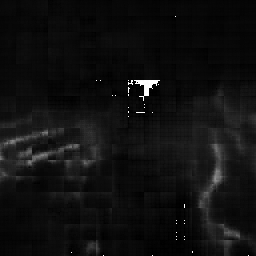}}\end{subfigure}
    \begin{subfigure}{2cm}{\includegraphics[width=2cm]{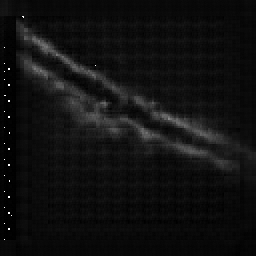}}\end{subfigure}
    \caption{}
    \end{subfigure}
     \caption{\label{fig:comparison} \footnotesize (a): $I_S.$ (b): Ground truth $I.$ (c): $\hat{I}$ generated using \gls{rsln}. (d) and (e): $\hat{I}$ using Pix2Pix with $g(I_S)$ and $g(I_{R,0}, I_S).$ (f): $\hat{I}$ using H-Nets with $g(I_{R,0}, I_S).$
    For Pix2Pix, we use the same generator architecture as in \gls{rsln}.
For H-Nets, we use an architecture with a similar number of parameters.
 }
\vspace{-2mm}
\end{figure*}

We implemented pipelines using Pix2Pix \cite{isola2017image} and harmonic networks (H-Nets) \cite{worrall2017harmonic,harmonic_pytorch}. 
Without a rotation selector to pre-infer the heading, the generator was pre-trained to condition on the radar image rotated with a coarse initial heading estimate $\theta_0,$ $g:(I_{R,0}, I_S) \rightarrow I,$ using a conditional GAN.
We also learn the deep embeddings $e_A$ and $e_B$ for pose computation, but rather search for the optimal $(x, y, \theta)$ from a 3D correlation volume.
The results are shown in \cref{tab:results} and \cref{fig:ratio_thresh}.
The use of harmonic networks resulted in small $e_\theta$ on the RobotCar Dataset, possibly because of the rotational equivariance property, but rather large $e_x$ and $e_y.$
For Pix2Pix, we also show examples of the synthetic radar image when the generator does not condition on radar data, $g:(I_S)\rightarrow I,$ which leads to inferior results than learning $g:(I_{R,0}, I_S) \rightarrow I,$ shown in \cref{fig:comparison}.

\begin{figure*}[h]%
    \centering
    \begin{subfigure}{\figsizeL}
    \centering
    \begin{subfigure}{\figsize}{\includegraphics[width=\figsize]{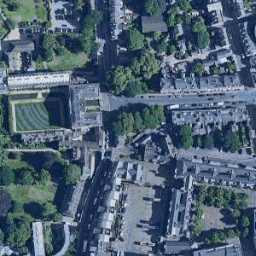}}\end{subfigure}
    \begin{subfigure}{\figsize}{\includegraphics[width=\figsize]{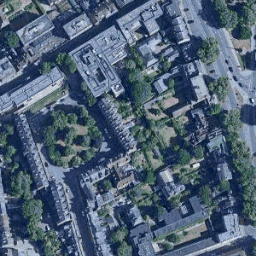}}\end{subfigure}
    \begin{subfigure}{\figsize}{\includegraphics[width=\figsize]{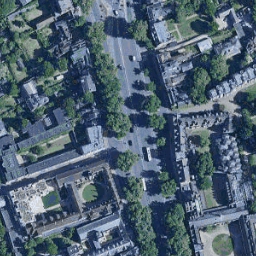}}\end{subfigure}
    \begin{subfigure}{\figsize}{\includegraphics[width=\figsize]{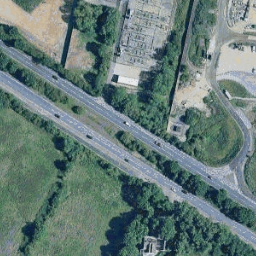}}\end{subfigure}
%    \begin{subfigure}{\figsize}{\includegraphics[width=\figsize]{figs/ral-new/sm_A6}}\end{subfigure}
    \caption{}
    \end{subfigure}
    \begin{subfigure}{\figsizeL}
    \centering
    \begin{subfigure}{\figsize}{\includegraphics[width=\figsize]{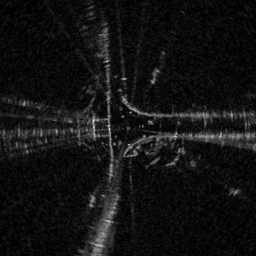}}\end{subfigure}
    \begin{subfigure}{\figsize}{\includegraphics[width=\figsize]{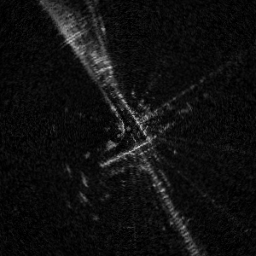}}\end{subfigure}
    \begin{subfigure}{\figsize}{\includegraphics[width=\figsize]{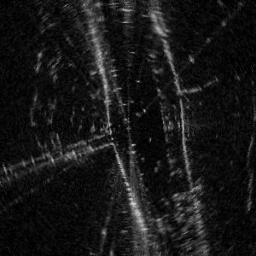}}\end{subfigure}
    \begin{subfigure}{\figsize}{\includegraphics[width=\figsize]{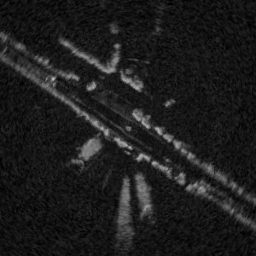}}\end{subfigure}
%    \begin{subfigure}{\figsize}{\includegraphics[width=\figsize]{figs/ral-new/sm_B6}}\end{subfigure}
    \caption{}
    \end{subfigure}
    \begin{subfigure}{\figsizeL}
    \centering
    \begin{subfigure}{\figsize}{\includegraphics[width=\figsize]{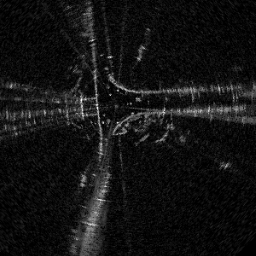}}\end{subfigure}
    \begin{subfigure}{\figsize}{\includegraphics[width=\figsize]{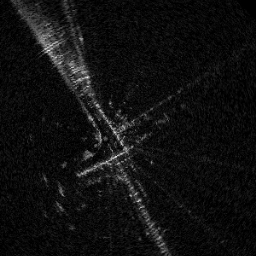}}\end{subfigure}
    \begin{subfigure}{\figsize}{\includegraphics[width=\figsize]{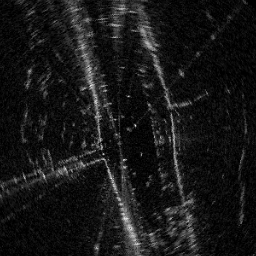}}\end{subfigure}
    \begin{subfigure}{\figsize}{\includegraphics[width=\figsize]{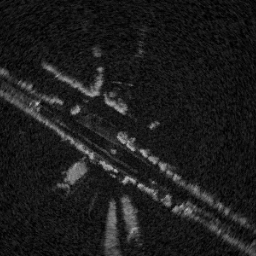}}\end{subfigure}
%    \begin{subfigure}{\figsize}{\includegraphics[width=\figsize]{figs/ral-new/sm_gt6}}\end{subfigure}
    \caption{}
    \end{subfigure}
    \begin{subfigure}{\figsizeL}
    \centering
    \begin{subfigure}{\figsize}{\includegraphics[width=\figsize]{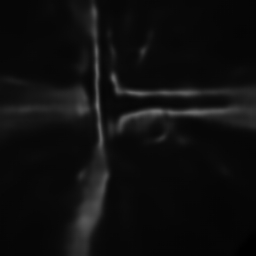}}\end{subfigure}
    \begin{subfigure}{\figsize}{\includegraphics[width=\figsize]{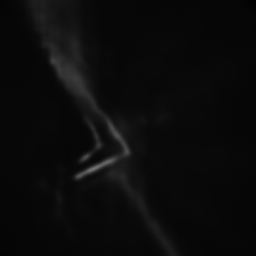}}\end{subfigure}
    \begin{subfigure}{\figsize}{\includegraphics[width=\figsize]{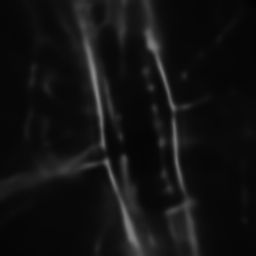}}\end{subfigure}
    \begin{subfigure}{\figsize}{\includegraphics[width=\figsize]{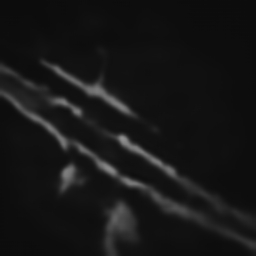}}\end{subfigure}
%    \begin{subfigure}{\figsize}{\includegraphics[width=\figsize]{figs/ral-new/sm_Ai6}}\end{subfigure}
    \caption{}
    \end{subfigure}
    \begin{subfigure}{\figsizeL}
    \centering
    \begin{subfigure}{\figsize}{\includegraphics[width=\figsize]{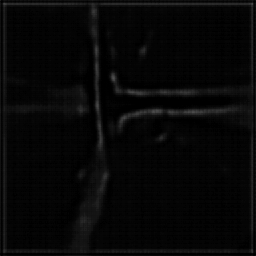}}\end{subfigure}
    \begin{subfigure}{\figsize}{\includegraphics[width=\figsize]{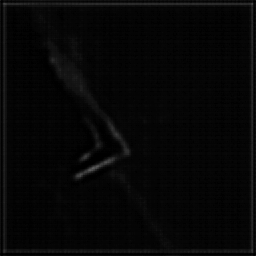}}\end{subfigure}
    \begin{subfigure}{\figsize}{\includegraphics[width=\figsize]{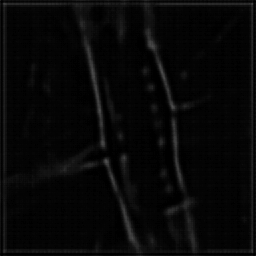}}\end{subfigure}
    \begin{subfigure}{\figsize}{\includegraphics[width=\figsize]{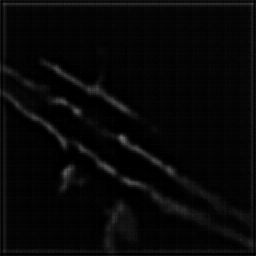}}\end{subfigure}
%    \begin{subfigure}{\figsize}{\includegraphics[width=\figsize]{figs/ral-new/sm_Am6_enhanced}}\end{subfigure}
    \caption{}
    \end{subfigure}
    \caption{\label{fig:further_quantitative} \footnotesize Additional qualitative results from the test set of the RobotCar Dataset (rows 1-3) and the Highway Dataset (row 4). (a): Satellite image $I_S.$ (b): Radar image with rotation offset uncovered, $\hat{I}_{R,*}.$ (c): Ground truth $I.$ (d): Synthetic radar image $\hat{I}.$(e):Deep embedding $e_A(\hat{I}).$
    \vspace{-4mm}
}

\end{figure*}

%As shown in \cref{fig:comparison}, the radar observes a limited and specific part of the scene covered by the satellite image. 
%Without conditioning using radar data, the synthetic images in (d) can appear different than what the radar actually measures, making pose estimation difficult.
%While the synthetic images are more realistic when conditioning on $I_{R,0},$ the results in (e) and (f) are still inferior comparing to (c), as the generator in \gls{rsln} utilises $\hat{I}_{R, *}$ which has approximately the same heading as the input satellite image.

As shown in \cref{fig:comparison}, without conditioning using radar data, the synthetic images in (d) can appear different than what the radar actually measures, making pose estimation difficult.
While they are more realistic when conditioning on $I_{R,0},$ the results in (e) and (f) are still inferior comparing to (c), as the generator in \gls{rsln} utilises $\hat{I}_{R, *}$ which has approximately the same heading as the input satellite image.

An experiment was performed where we modified the generator in RSL-Net to learn $g(I_S)\rightarrow I$ instead, and kept other modules the same, to further demonstrate the added benefit of conditioning the generator using live radar data.

%To demonstrate the advantage of our correlation-based search for finding the translation offset over regression, we trained a network that takes in $\hat{I}_{R, *}$ and $\hat{I}$ as inputs, and directly regresses the $(x, y)$ offset, where the last layer is a fully connected layer.
%We have also benchmarked against a baseline \gls{stn} \cite{jaderberg2015spatial} method, where given $\hat{I}_{R, *}$ and $\hat{I},$ the network regresses the $(x, y)$ offset, and transforms $\hat{I}_{R, *}$ to register onto $\hat{I}.$
%While both methods regress $(x, y),$ the first regression method is supervised by a loss on pose and the \gls{stn} method is supervised with an L1 loss on the registered images.
%The results are shown in \cref{tab:results}. 
%To ensure a fair comparison, the rotation is pre-inferred using the rotation selector as in \gls{rsln} in order to generate $\hat{I}$ correctly.

To demonstrate the advantage of our correlation-based search for finding the translation offset over regression, we trained a network that takes in $\hat{I}_{R, *}$ and $\hat{I}$ as inputs, and directly regresses the $(x, y)$ offset, where the last layer is a fully connected layer.
We have also benchmarked against a baseline \gls{stn} \cite{jaderberg2015spatial} method, where given $\hat{I}_{R, *}$ and $\hat{I},$ the network regresses the $(x, y)$ offset, and transforms $\hat{I}_{R, *}$ to register onto $\hat{I}.$
Both methods regress $(x, y);$ the former is supervised by a loss on pose while the latter is supervised with an L1 loss on the registered images.
The results are shown in \cref{tab:results}. 
To ensure a fair comparison, the rotation is pre-inferred using the rotation selector as in \gls{rsln} in order to generate $\hat{I}$ correctly.

We have also experimented with other simpler methods, such as directly regressing $(x, y, \theta)$ given the input satellite-radar pair, or finding the pose using Canny edges extracted.
%These methods led to poor results, as radar and satellite images are drastically different forms of data, and requires careful treatment prior to pose computation.
These methods led to poor results, as shown in Table \ref{tab:results}.
We believe this is due to radar and satellite images being drastically different forms of data, and requires careful treatment prior to pose computation.

Additional examples of images at various stages of RSL-Net are shown in \cref{fig:further_quantitative}.

%===============================================================================
\section{Conclusion}
In this work, we introduce a novel framework to learn the metric localisation of a ground range sensor using overhead imagery, namely \gls{rsln}. 
From a coarse initial pose estimate, \gls{rsln} computes a refined metric localisation by comparing live radar images against Google satellite images.
\gls{rsln} allows a vehicle to localise in real-time using an on-board radar without relying on any prior radar data, offering a fallback solution when the vehicle fails to localise against prior sensor maps.
The experimental validation demonstrates the capability for \gls{rsln} to perform even outside of well-structured, urban environments.

\bibliographystyle{IEEEtran}
\bibliography{main_rnr}

\end{document}